\documentclass{article}
\usepackage{ijcai03}
\usepackage{times}  
\usepackage{latexsym}
\usepackage{epsfig}

\title{Bayesian Information Extraction Network}
\author{Leonid Peshkin and Avi Pfeffer \\
 Harvard University \\ 
 Cambridge, MA 02138, USA \\
 \sf{\{pesha,avi\}@eecs.harvard.edu}}

\begin{document}

\maketitle

\begin{abstract}
Dynamic Bayesian networks (DBNs) offer an elegant way to integrate various
aspects of language in one model. Many existing algorithms developed for
learning and inference in DBNs are applicable to probabilistic language
modeling. To demonstrate the potential of DBNs for natural language
processing, we employ a DBN in an information extraction task. We show how to
assemble wealth of emerging linguistic instruments for shallow parsing,
syntactic and semantic tagging, morphological decomposition, named entity
recognition etc. in order to incrementally build a robust information
extraction system. Our method outperforms previously published results on an
established benchmark domain.

\end{abstract}

\section{Information Extraction} 

Information extraction ({\sc ie}) is the task of filling in template
information from previously unseen text which belongs to a pre-defined
domain. The resulting database is suited for formal queries and
filtering. {\sc ie} systems generally work by detecting patterns
in the text that help identify significant information. 
Researchers have shown~\cite{Freitag99,Ray01} that a probabilistic
approach allows the construction of robust and well-performing systems.
However, the existing probabilistic systems are generally based on
Hidden Markov Models ({\sc hmm}s). Due to this relatively impoverished
representation, they are unable to take advantage of the wide array of
linguistic information used by many non-probabilistic {\sc ie} systems.
In addition, existing {\sc hmm}-based systems model each
target category separately, failing to capture relational
information, such as typical target order, or the fact that each 
element only belongs to a single category.
This paper shows how to incorporate a wide array of knowledge into a
probabilistic {\sc ie} system, based on dynamic Bayesian networks ({\sc
dbn})---a rich probabilistic representation that generalizes {\sc
hmm}s. 

Let us illustrate {\sc ie} by describing seminar
announcements which got 
established as one of the most popular benchmark domains in the
field~\cite{Califf99,Freitag99,Soderland99,Roth01,Ciravegna01}. People
receive dozens of seminar announcements weekly and need to manually extract
information and paste it into personal organizers. The goal of an {\sc ie}
system is to automatically identify target fields such as location and
topic of a seminar, date and starting time, ending time and speaker.
Announcements come in many formats, but usually follow some pattern. We often
find a header with a gist in the form ``{\tt PostedBy: john@host.domain; 
 Who: Dr. Steals; When: 1 am;}'' and so forth. Also in the
body of the message, the speaker usually precedes both location and starting
time, which in turn precedes ending time as in: {\tt ``Dr. Steals presents in
Dean Hall at one am.''} The task is complicated since some fields may be
missing or may contain multiple values.


This kind of data falls into the so-called semi-structured text category.
Instances obey certain structure and usually contain information for most of
the expected fields in some order. There are two other categories: free text
and structured text.
In {\em structured text}, the positions of the information fields are fixed
and values are limited to pre-defined set. Consequently, the {\sc ie} systems
focus on specifying the delimiters and order associated with each field. 
At the opposite end lies the task of extracting information from {\em free
text} which, although unstructured, is assumed to be
grammatical. Here {\sc ie} systems rely more on syntactic, semantic and
discourse knowledge in order to assemble relevant information potentially
scattered all over a large document.

\begin{table*}[bt]
\begin{center}
\begin{tabular}{c||c|c|c|c|c|c|c|c|c|c}
Position & 1 & 2 & 3 & 4 & 5 & 6 & 7 & 8 & 9 & 10 \\
\hline \bf
 Tag &\multicolumn{2}{c|}{$<$speaker$>$}& & &\multicolumn{2}{c|}{$<$location$>$}& &\multicolumn{2}{c|}{$<$s-time$>$}& \\
Phrase&{\it Doctor}&{\it Steals}&{\it Presents}&{\it in}&{\it Dean}&{\it Hall}&{\it at}  & 1   &{\it am}& . \\
\hline
 Lemma  & Dr.  &      & present  & in   &     & hall & at  &   &{\bf am}&  \\
 PoS & {\sc nnp}&{\sc nnp(vb)}&{\sc vb(nns)}&{\sc in}&{\sc nnp}&{\sc
 nn(nnp)}&{\sc in}&{\sc cd}&{\sc nn(rb)}& . \\
Syn.segm.& NP &NP ({\sc vp}) & VP & PP & NP & NP & PP & NP  &NP({\sc vp})& \\
Semantic &{\bf Title}&LstName&        &      &LstName &Location&   &   &Time&  \\
 Length  &    3 & 6      &   8      &   2  & 4    & 4    & 2  & 1    & 2 & 1 \\
 Case    &Upper & Upper  &  Upper   &lower & Upper&Upper&lower&      &lower&  \\
\end{tabular}
\caption{Sample phrase and its representation in multiple feature values for
 ten tokens.}
\label{example}
\end{center}
\end{table*}

{\sc ie} algorithms face different challenges depending on the extraction
targets and the kind of the text they are embedded in. In some cases, the
target is uniquely identifiable (single-slot), while in others, the targets
are linked together in multi-slot association frames. For example, a
conference schedule has several slots for related speaker, topic and time of
the presentation, while a seminar announcement usually refers to a unique
event. Sometimes it is necessary to identify each word in a target slot,
while some benefit may be reaped from partial identification of the target,
such as labeling the beginning or end of the slot separately. 
Many applications involve processing of domain-specific jargon like {\it
Internetese}---a style of writing prevalent in news groups, e-mail messages,
bulletin boards and online chat rooms. Such documents do not follow a good
grammar, spelling or literary style. Often these are more like a
stream-of-consciousness ranting in which ascii-art and pseudo-graphic
sketches are used and emphasis is provided by all-capitals, or
using multiple exclamation signs. As we exemplify below, syntactic analysers
easily fail on such corpora.

Other examples of {\sc ie} application domains include job
advertisements~\cite{Califf99} ({\sc rapier}), executive
succession~\cite{Soderland99} ({\sc whisk}), restaurant
guides~\cite{Muslea01} ({\sc stalker}), biological publications~\cite{Ray01}
etc. Initial interest in the subject was stimulated by ARPA's Message
Understanding Conferences (MUC) which put forth challenges e.g. parsing
newswire articles related to terrorism (see e.g.
Mikheev~\shortcite{Mikheev98}). Below we briefly review various {\sc ie}
systems and approaches which mostly originated from MUC competitions.

Successful {\sc ie} involves identifying abstract patterns in the way
information is presented in text. Consequently, all previous work necessarily
relies on some set of textual features. The overwhelming majority of existing
algorithms operate by building and pruning sets of induction rules defined on
these features ({\sc srv, rapier, whisk, lp}$^2$).
%
There are many features that are potentially helpful for extracting specific
fields, e.g. there are tokens and delimiters that signal the
beginning and end of particular types of information. Consider an example in
table~\ref{example} which shows how the phrase ``{\tt Doctor Steals presents
in Dean Hall at one am.}'' is represented through feature values.
For example, the
lemma ``am'' designates the end of a time field, while the semantic feature
``Title'' signals the speaker, and the syntactic category {\sc nnp} (proper
noun) often corresponds to speaker or location. Since many researchers use
the seminar announcements domain as a testbed, we have chosen this domain in
order to have a good basis of comparison.
%

One of the systems we compare to (specifically designed for single-slot
problems) is SRV~\cite{Freitag98}. It is built on three classifiers of
text fragments. The first classifier is a simple look-up table containing all
correct slot-fillers encountered in the training set. The second one computes
the estimated probability of finding the fragment tokens in a correct
slot-filler. The last one uses constraints obtained by rule induction over
predicates like token identity, word length and capitalization, and simple
semantic features.

{\sc rapier}~\cite{Califf99} is fully based on bottom-up rule induction on
the target fragment and a few tokens from its neighborhood. The rules are
templates specifying a list of surrounding items to be matched and
potentially, a maximal number of tokens for each slot. Rule generation begins
with the most specific rules matching a slot. Then rules for identical slots
are generalized via pair-wise merging, until no improvement can be made.
Rules in {\sc rapier} are formulated as lexical and semantic constraints and
may include {\sc PoS} tags.
{\sc whisk}~\cite{Soderland99} uses constraints similar to {\sc rapier},
but its rules are formulated as regular expressions with wild cards for
intervening tokens. Thus, {\sc whisk} encodes a 
relative, rather than absolute position of tokens with respect to the target.
This enables modeling long distance dependencies in the text.
{\sc whisk} performs well on both single-slot and multi-slot extraction tasks.

Ciravegna~\shortcite{Ciravegna01} presents yet another rule induction method
({\sc lp})$^2$. He considers several candidate features such as lemma,
lexical and semantic categories and capitalization to form a set of rules for
inserting tags into text. Unlike other approaches, ({\sc lp})$^2$ generates
separate rules targeting the beginning and ending of each slot. This allows
for more flexibility in subjecting partially correct extractions to several
refinement stages, also relying on rule induction to introduce corrections.
%
Emphasizing the relational aspect of the domain, Roth and
Yoh~\shortcite{Roth01} developed a knowledge representation language that
enables efficient feature generation. They used the features in a multi-class
classifier {\sc snow-ie} to obtain the desired set of tags. The resulting
method ({\sc snow-ie}) works in two stages: the first filters out the
irrelevant parts of text, while the second identifies the relevant slots.

Freitag \& McCallum~\shortcite{Freitag99} use hidden Markov models ({\sc hmm}). A
separate {\sc hmm} is used for each target slot. No preprocessing or features
is used except for the token identity. 
For each hidden state, there is a probability distribution over
tokens encountered as slot-fillers in the training data. 
Weakly analogous to templates, hidden state transitions
encode regularities in the slot context. In particular prefix and suffix
states are used in addition to target and background slots to capture words
frequently found in the neighborhood of targets. Ray\&Craven~\shortcite{Ray01}
make one step further by setting {\sc hmm} hidden states in a product space of
syntactic chunks and target tags to model the text structure.
The success of the {\sc hmm}-based approaches demonstrate the viability of
probabilistic methods for this domain.  However, they do not take
advantage of the linguistic information used by the other approaches.
Furthermore, they are limited by using a separate {\sc hmm} for each target
slot, rather than extracting data in an integrated way.

The main contribution of this paper is in demonstrating how to integrate
various aspects of language in a single probabilistic model, to incrementally
build a robust information extraction system based on a Bayesian network.
This system overcomes the following dilemma. It is tempting to use a lot of
linguistic features in order to account for multiple aspects of text
structure. However, deterministic rule induction approaches seem vulnerable
to the performance of feature extractors in pre-processing steps. This
presents a problem since syntactic instruments that have been trained on
highly-polished grammatical corpora, are particularly unreliable on weakly
grammatical semi-structured text. Furthermore incorporating many features
complicates the model which often has to be learned from sparse data, which
harms performance of classifier-based systems.

\section{Features}
\label{features}

Our approach is statistical, which generally speaking means that learning
corresponds to inferring frequencies of events. The statistics we collect
originates in various sources. Some statistics reflect regularities
of the language itself, while others correspond to the peculiarities of the
domain. With this in mind we design features which reflect both aspects.
There is no limitation on the possible set of features. 
Local features like part-of-speech, number of characters in the
token, capitalization and membership in syntactic phrase are quite
customary in the {\sc ie}. In addition one could obtain such characteristics
of the word as imagibility, frequency of use, familiarity, or even predicates
on numerical values.
Since there is no need for features to be local, one might find useful 
including frequency of a word in the training corpus or number of
occurrences in the document. 
Notice that the same set of features would work for many domains.
This includes semantic features along with orthographic and syntactic
features.

Before we move on to presenting our system for probabilistic reasoning,
let us discuss in some detail notation and methods we used in
preliminary data processing and feature extraction.
To use the data efficiently, we need to {\em factor} the text into
``orthogonal'' features. Rather than working with thousands of listems
(generic words\footnote{A {\em word} is a sequence of
alphabetical characters, which has some meaning assigned to it. This would
cover words found in general and special vocabulary as well abbreviations,
proper names and such.}) in the vocabulary, and combining their features, we
compress the vocabulary by an order of magnitude by lemmatisation or
stemming. Orthographic and syntactic information is kept in feature variables
with just a few values each.

\subsubsection{Tokenization} 
Tokenization is the first step of textual data processing. A token is a minimal
part of text which is treated as a unit in subsequent steps. In our case
tokenization mostly involves separating punctuation characters from words.
This is particularly non-trivial for separating a period~\cite{Manning} since
it requires identifying sentence boundaries. Consider a sentence: {\tt Speaker: Dr.
Steals, Chief Exec. of rich.com, worth $\$ 10.5$ mil. } 

\subsubsection{Lemmatisation}
We have developed a simple lemmatiser which combines outcome of some standard
lematisers and stemmers into a look-up table. Combined with lemmatisation is
a step of spell checking to catch misspelled words. This is done by
interfacing with the {\sc unix} {\em ispell} utility.

\subsubsection{Gazetteer}
Our original corpus contains about 11,000 different listems. This does not
take into account tokens consisting of punctuation characters, numbers and
such. About 10\% are proper nouns. The question of building a vocabulary
automatically was previously addressed in {\sc ie} literature(see e.g.
Riloff~\shortcite{Riloff96}). We use the intersection of two sets. The first set
consists of words encountered as part of target fields and in their
neighborhood. The second set consists of words frequently seen in the
corpus. 
Aside from vocabulary there are two reserved values for Out-of-Vocabulary
(OoV) words and Not-a-Word (NaW). For example see blank slots in the {\em
lemma} row of Table~\ref{example}. 
The first category encodes rare and
unfamiliar words, which are still identified as words according to their part
of speech. The second category is for mixed alpha-numerical tokens,
punctuation and symbolic tokens. 

\subsubsection{Syntactic Categories} 
We used {\tt LTChunk} software from U.of Edinburgh NLP
group~\cite{Mikheev98}. It produces $47$ PoS tags from UPenn TreeBank
set~\cite{PennTreebank}. We have clustered these into 7 categories: 
cardinal numbers ({\sc cd}), nouns ({\sc nn}), proper nouns ({\sc nnp}),
verbs ({\sc vb}), punctuation (.), preposition/conjunction ({\sc in}) and
other ({\sc sym}).  The choice of clusters seriously influences the
performance, while keeping all $47$ tags will lead to large CPTs and sparse
data. 

\subsubsection{Syntactic Chunking} 
Following Ray\&Craven~\shortcite{Ray01}, we obtain syntactic segments (aka
syntactic chunks) by running the {\em Sundance} system~\cite{Riloff96} and
flattening the output into four categories corresponding to noun phrase 
({\sc np}), verb phrase ({\sc vp}), prepositional phrase ({\sc pp}) and 
other ({\sc n/a}).
Table~\ref{example} shows a sample outcome. Note that both the
part-of-speech tagger and the syntactic chunker easily get confused by
non-standard capitalization of a word ``Presents'' as shown by incorrect
labels in parenthesis. ``Steals'' is incorrectly identified as a verb, whose
subject is ``Doctor'' and object is ``Presents''. Remarkably, other
state-of-the-art syntactic analysis tools~\cite{Charniak99,Ratnaparkhi99}
also failed on this problem. 

\subsubsection{Capitalization and Length}
Simple features like capitalization and length of word are used by many
researchers (e.g. SRV~\cite{Freitag99})
Case representation process is straightforward except for the choice of
number of categories. We found useful introducing an extra category for words
which contain both lower and upper case letters (not counting the initial
capital letter) which tend to be abbreviations. 

\subsubsection{Semantic Features} 
There are several semantic features which play important role in a variety of
application domains. In particular, it is useful to be able to recognize what
could be a person's name, geographic location, various parts of address, etc.
For example, we are using a list of secondary location identifiers provided
by US postal service, which identifies as such words like {\em hall, wing,
floor} and {\em auditorium}. We also use a list of $100000$ most popular names
from US census bureau; the list is augmented by rank which helps to decide in
favor of first or last name for cases like ``{\tt Alexander}''. In general this
task could be helped by using a hypernym feature of
WordNet project~\cite{WordNet}. The next section presents probabilistic model
which makes use of the aforementioned feature variables.

\section{BIEN}

We convert the {\sc ie} problem into a classification problem by assuming that
each token in the document belongs to a target {\em class} corresponding to
ether one of the target tags or the background (compare to
Freitag~\shortcite{Freitag99}). Furthermore, it seems important not to ignore
the information about interdependencies of target fields and document
segments. To combine advantages of stochastic models with feature-based
reasoning, we use a Bayesian network.

A dynamic Bayesian network ({\sc dbn}) is ideal for representing probabilistic
information about these features. Just like a Bayesian network, it encodes
interdependence among various features. In
addition, it incorporates the element of time, like an {\sc hmm}, so that
time-dependent patterns such as common orders of fields can be represented.
All this is done in a compact representation that can be learned from data.
We refer to a recent dissertation~\cite{MurphyPhD} for a
good overview of all aspects of Dynamic Bayesian Networks.

\begin{figure}[hbt]
\centerline{\epsfig{file=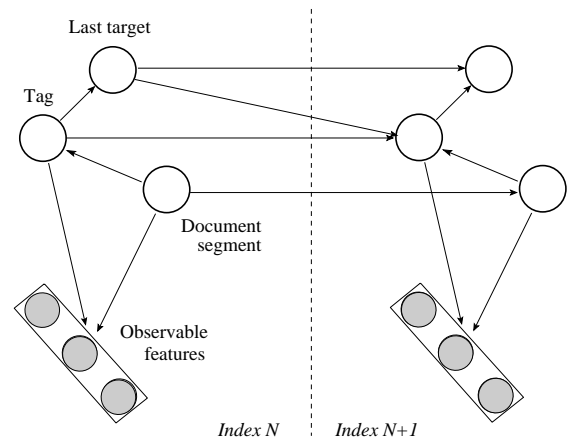,width=2.9in}}
\caption{A schematic representation of {\sc bien}.}
\label{2tbn}
\end{figure}

Each document is considered to be a single stream of tokens. In our {\sc dbn},
called the Bayesian Information Extraction Network ({\sc bien}), the same
structure is repeated for every index. Figure~\ref{2tbn} presents the
structure of {\sc bien}. This structure contains state variables
and feature variables. The most important state variable, for our purposes,
is ``Tag'' which corresponds to information we are trying to extract.
This variable classifies each token according to its target information
field, or has the value ``background'' if the token does not belong to any
field. ``Last Target'' is another hidden variable which reflects the order in
which target information is found in the document. This variable is our way
of implementing a memory in a ``memory-less'' Markov model. Its value is
deterministically defined by the last non-background value of ``Tag''
variable. Another hidden variable, ``Document Segment'', is introduced to
account for differences in patterns between the header and the main body of
the document. The former is close to the structured text format, while the
latter to the free text. ``Document Segment'' influences ``Tag'' and together these
two influence the set of observable variables which represent features of the
text discussed in section~\ref{features}.
Standard inference algorithms for {\sc dbn}s are similar to those for {\sc hmm}s.
In a {\sc dbn}, some of the variables will typically be observed, while
others will be hidden.  
The typical inference task is to determine the probability
distribution over the states of a hidden variable over time, given
time series data of the observed variables.
This is usually accomplished using the forward-backward algorithm.
Alternatively, we might want to know the most likely sequence of
hidden variables. This is accomplished using the Viterbi algorithm.
Learning the parameters of a {\sc dbn} from data is accomplished using the
{\sc em} algorithm (see e.g. Murphy~\shortcite{MurphyPhD}).
Note that in principle, parts of the system could be trained separately on
independent corpus to improve performance. For example, one could learn
independently the conditional vocabulary of email/newsgroup headers, or learn a
probability of part-of-speech conditioned on a word, to avoid dependence on
external PoS taggers. Also prior knowledge about the domain and
the language could be set in the system this way. The fact that {\em etime}
almost never
precedes {\em stime} as well as the fact that {\em speaker} is never a verb
could be encoded in a conditional probability table (CPT).
In large {\sc dbn}s, exact inference algorithms are intractable, and so a
variety of approximate methods have been developed.  However,
the number of hidden state variables in our model is small enough to allow
exact algorithms to work. Indeed, all hidden nodes in our model are
discrete variables which assume just a few values. ``DocumentSegment'' is
binary in \{{\em Header}, {\em Body}\} range; ``LastTarget'' has as many
values as ``Tag''---four per number of target fields plus one for the
background. 



\section{Results}

Several researchers have reported results on the CMU seminar announcements
corpus, which we have chosen in order to have a good basis of comparison.
The CMU seminar announcements corpus consists of 485 documents. Each
announcement contains some tags for target slots. On average {\em starting
time} 
appears twice per document, while {\em location} and {\em speaker} $1.5$ times, with up to
$9$ {\em speaker} slots and $4$ {\em location} slots per document. Sometimes multiple
instances of the same slot differ, e.g. speaker {\tt Dr.~Steals} also appears 
as {\tt Joe Steals}\footnote{Obtaining $100\%$ performance
on the original corpus is impossible since some tags are misplaced and in
general the corpus is not marked uniformly---sometimes secondary occurrences
are ignored.}. {\em Ending time}, {\em speaker} and {\em location} are missing from $48\%$, $16\%$
and $5\%$ of documents correspondingly.
In order to demonstrate our method, we have developed a web site which works
with arbitrary seminar announcement and reveals some semantic tagging. We
also make available a list of errors in the original corpus, along with our
new derivative seminar announcement corpus\footnote{The corpus and demo for this paper are available from {\tt http://www.eecs.harvard.edu/\~{}pesha/papers.html}}.


\begin{table}[!t]
\begin{center}
\begin{tabular}{l||c|c|c|c}
System            & stime   & etime   &location & speaker \\
\hline 
{\sc snow-ie} &{\bf99.6}& 96.3    & 75.2    & 73.8 \\
{\sc rapier}  & 95.9    & 94.6    & 73.4    & 53.1 \\
{\sc srv}     & 98.5    & 77.9    & 72.7    & 56.3 \\   
{\sc hmm}     & 98.5    & 62.1    & 78.6    & 76.6 \\
{\sc whisk}   & 92.6    & 86.1    & 66.6    & 18.3 \\
{\sc(lp)}$^2$ & 99.0    & 95.5    & 75.0    &{\bf77.6}\\
\hline 
{\sc bien}    & 96.0    &{\bf98.8}&{\bf87.1}& 76.9 \\
\end{tabular}
\caption{F1 performance measure for various {\sc ie} systems.}
\label{results}
\end{center}
\end{table}

\begin{table}[!b]
\begin{center}
\begin{tabular}{l||c|c|c|c}
{\sc bien}  & stime   & etime   &location & speaker \\
\hline 
no semantic  & 92.0   & 95.0    & 73.0    & 57.0 \\
no memory    & 99.0   & 95.5    & 75.0    & 74.6 \\
no lemma     & 54.5   & 53.4    & 29.0    & 18.4 \\
no length    & 98.1   & 98.6    & 87.9    & 79.8 \\
no case      & 98.1   & 98.5    & 78.3    & 71.8 \\
\hline 
complete     & 98.0   & 98.8    & 88.5    & 79.7 \\
\end{tabular}
\caption{F1 performance comparison across implementations of {\sc bien} with
disabled features.}
\label{results2}
\end{center}
\end{table}

The performance is calculated in the usual way, by {\em precision}
$P=\frac{\mbox{\rm correct answers}}{\mbox{\rm answers produced}}$
and 
{\em recall} $R=\frac{\mbox{\rm correct answers}}{\mbox{\rm total correct}}$
combined into $F$ measure geometrical average $F=\frac{2 P \cdot R}{P + R}$. 
We report 
results using the same ten-fold cross validation test as 
other publications concerning this data set~\cite{Roth01,Ciravegna01}. The
data is split randomly into training and testing set. The reported results
are averaged over five runs.
Table~\ref{results} presents a comparison with numerous previous attempts at
the CMU seminar corpus. The figures are taken from Roth and Yih~\shortcite{Roth01}.
{\sc bien} performs comparably to the best system in each category, while
notably outperforming other systems in finding {\em location}.  This is
partly due to the ``LastTarget'' variable.
%
"LastTarget" variable turns out to be generally useful. Here is the learned
conditional probability table (CPT) for $P(Target|LastTarget)$, where the element
$(I,J)$ corresponds to the probability to get target tag $J$ after target
tag $I$ was seen. We learn that initial tag is {\em stime} or {\em speaker}
with 2:1 likelihood ratio; {\em etime} is naturally the most likely follower to
{\em stime} and in turn forecasts {\em location}.\vspace{3mm}\\
Last Target
\begin{tabular}{l|cccc}
             & \multicolumn{4}{c}{Current Tag} \\
          &stime &etime&location & speaker \\
\hline 
  none    &{\bf.66}& 0   &  .01 & .33 \\
  stime   & .07    &{\bf .46}&  .32 & .14 \\
  etime   & .19    &.002 &{\bf .63} & .18 \\
location  & .37    &.004 & .02  &{\bf.61} \\
 speaker  &{\bf.64}&.002 & .13  & .22 \\
\end{tabular} \\

Other variables turn out to be useless, e.g. the number of characters
does not add anything to the performance, and neither does the initially
introduced ``SeenTag'' variable which kept track of all tags seen up to the
current position. Table~\ref{results2} presents performance of {\sc bien}
with various individual features turned off. Note that figures for complete
{\sc bien} are slightly better than in Table~\ref{results} since we pushed
the fraction of the training data to the maximum.
Capitalization helps identify
{\em location} and {\em speaker}, while losing it does not damage performance
drastically. Although information is reflected in syntactic and semantic
features, most names in documents do not identify a speaker. One would hope
to capture all relevant information by syntactic and semantic categories,
however {\sc bien} does not fare well without observing ``Lemma''.
Losing the semantic feature seriously undermines performance in {\em location}
and {\em speaker} categories---ability to recognize names is
rather valuable for many domains.

\begin{figure}[!t]
\centerline{\epsfig{file=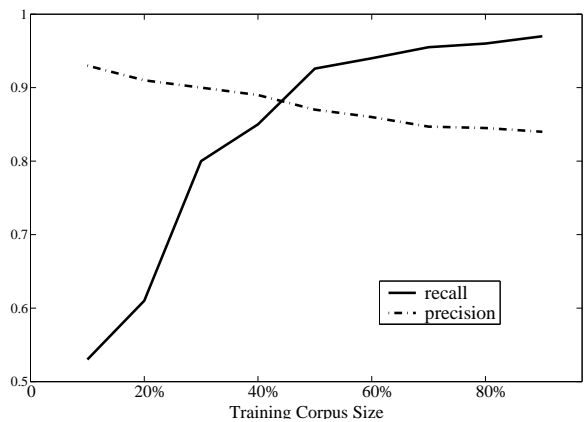,width=3.02in,angle=0}}
\caption{A learning curve for precision and recall with growing training
sample size.}
\label{learncur}
\end{figure}

Reported figures
are based on 80\%-20\% split of the corpus. Increasing the size of training
corpus did not dramatically improve the performance in terms of $F$ measure,
as further illustrated 
by Figure~\ref{learncur}, which presents a learning curve---precision and
recall averaged over all fields, as a function of training data fraction.
Trained on a small sample, {\sc bien} acts very conservatively rarely picking
fields, therefore scoring high precision and poor recall. Having seen
hundreds of target field instances and tens of thousands of negative samples,
{\sc bien} learns to generalize, which leads to generous tagging i.e. lower
precision and higher recall.

So far we provide results obtained on the original CMU seminar
announcements data, which is not very challenging. Most documents
contain the header section with all the target fields easily identifiable
right after the corresponding key word. We have created a derivative dataset
in which documents are stripped of headers and two extra fields are sought:
{\em date} and {\em topic}. Indeed this corpus
turned out to be more difficult, with our current set of features 
we obtain only $64\%$ performance on {\em speaker} and $68\%$ performance on 
{\em topic}. {\em Date} does not present a challenge except for cases of
regular weekly events or relative dates like ``tomorrow''.
Admittedly, the bootstrapping test performance is not a guarantee of systems
performance on novel data since preliminary processing, i.e. tokenization and
gazetteering, as well as choice for PoS tag set, lead to a strong bias towards
the training corpus.

\section{Discussion}

We have described how to integrate various aspects of language into a single
probabilistic model, and to incrementally build a robust IE system based on a
Bayesian network. Currently, we are working on 
learning the structure of {\sc bien} automatically. It seems to subject itself
nicely to structural {\sc em}~\cite{Friedman98,MurphyPhD}. The first step is
automatic selection of relevant features. 
Another direction of current work is using approximate inference. We have
tried LBP (Loopy-belief Propagation)~\cite{MurphyUAI99,MurphyPhD}, but for
the current structure of {\sc bien} it seems to give no gain. More challenging
applications which require larger, stronger connected networks, will benefit
from approximate inference algorithms. It will enable quick on-line inference
on the network learned off-line with exact methods, as well as learning for
cases where exact inference is infeasible.
One such network will result from integrating a PoS tagger and other feature
extractors into {\sc bien}. This is a natural extension of {\sc bien} since
various text processing routines are mutually dependent. Consider for example  
PoS tagging, sentence boundary detection and named entities recognition. 
Another complex {\sc bien} structure will result if we try
to better reflect complex relational information~\cite{Califf99,Roth01,Roth02} e.g.
to process cases like seminar cancellations and rescheduling; and handle
multi-slot extraction, e.g. multiple seminar announcements and conference
schedules.

\section*{Acknowledgments}
Kevin Murphy provided {\it BNT}, Kobi Gal helped to handle the corpus,
anonymous reviewers gave helpful feedback. 




\nocite{McCallum00}

\bibliography{/home/lair/pesha/BiB/headers,/home/lair/pesha/BiB/BIEN}
\bibliographystyle{named}

\end{document}